\documentclass[conference]{IEEEtran}
\usepackage{times}

\usepackage[numbers]{natbib}
\usepackage{multicol}
\usepackage[bookmarks=true]{hyperref}
\usepackage{amsfonts}
\usepackage{amsmath}
\usepackage{graphicx}
\usepackage{xcolor}
\usepackage{algorithmic}
\usepackage[linesnumbered,ruled,vlined]{algorithm2e}
\DeclareMathOperator*{\argmax}{arg\,max}
\usepackage{float}
\usepackage{booktabs}
\usepackage{bm}
\usepackage{makecell}
\usepackage{wrapfig}
\usepackage{bbding}  
\usepackage{pifont}  
\usepackage{amssymb}  
\usepackage{multirow}
\usepackage{xspace}

\begin{document}

\title{Towards Uncertainty Unification: \\A Case Study for Preference Learning}

\author{Shaoting Peng, Haonan Chen, Katherine Driggs-Campbell \\ University of Illinois Urbana-Champaign}

\maketitle

\begin{abstract}

Learning human preferences is essential for human-robot interaction, as it enables robots to adapt their behaviors to align with human expectations and goals. However, the inherent uncertainties in both human behavior and robotic systems make preference learning a challenging task. While probabilistic robotics algorithms offer uncertainty quantification, the integration of human preference uncertainty remains underexplored. To bridge this gap, we introduce uncertainty unification and propose a novel framework, uncertainty-unified preference learning (UUPL), which enhances Gaussian Process (GP)-based preference learning by unifying human and robot uncertainties. Specifically, UUPL includes a human preference uncertainty model that improves GP posterior mean estimation, and an uncertainty-weighted Gaussian Mixture Model (GMM) that enhances GP predictive variance accuracy. Additionally, we design a user-specific calibration process to align uncertainty representations across users, ensuring consistency and reliability in the model performance. Comprehensive experiments and user studies demonstrate that UUPL achieves state-of-the-art performance in both prediction accuracy and user rating. An ablation study further validates the effectiveness of human uncertainty model and uncertainty-weighted GMM of UUPL. Video and code are available at \href{https://sites.google.com/view/uupl-rss25/home}{https://sites.google.com/view/uupl-rss25/home}.
\end{abstract}

\IEEEpeerreviewmaketitle

\section{Introduction}

Uncertainty is pervasive in human-robot interaction (HRI), stemming from both robotic and human sources. On the robot side, uncertainties may arise from noisy sensor measurements, imperfect control models, unpredictable human behavior, and many other aspects~\cite{battula2024uncertainty}. Properly modeling these uncertainties enhances HRI efficiency and safety, while neglecting them can lead to significant risks \cite{Senanayake2024TheRO}. Established probabilistic techniques, such as the Kalman filter and partially observable Markov decision processes (POMDP), as well as modern methods leveraging large language models for uncertainty estimation \cite{Ren2023RobotsTA, Ren2024ExploreUC}, enable effective quantification and management of robot uncertainty, improving system performance.

In contrast, human uncertainty is inherently more challenging to quantify due to the stochastic nature of human decision-making. Psychological and cognitive science theories, such as signal detection theory~\cite{macmillan2002signal}, provide valuable frameworks for modeling this uncertainty. However, when humans interact with autonomous agents, their uncertainty becomes even more nuanced, shaped by subjective factors like trust, familiarity with the technology, and individual risk aversion~\cite{FRANK2024102732}. Additionally, dynamic interactions with robots during HRI tasks introduce more variability, as users adapt their behaviors based on perceived system performance \cite{Guo2020ModelingAP}. Despite these complexities, accounting for human uncertainty is crucial for optimizing both user experience and system efficacy, and further facilitating smoother collaboration.

\begin{figure}[!t]
    \centering
    \includegraphics[width=\columnwidth]{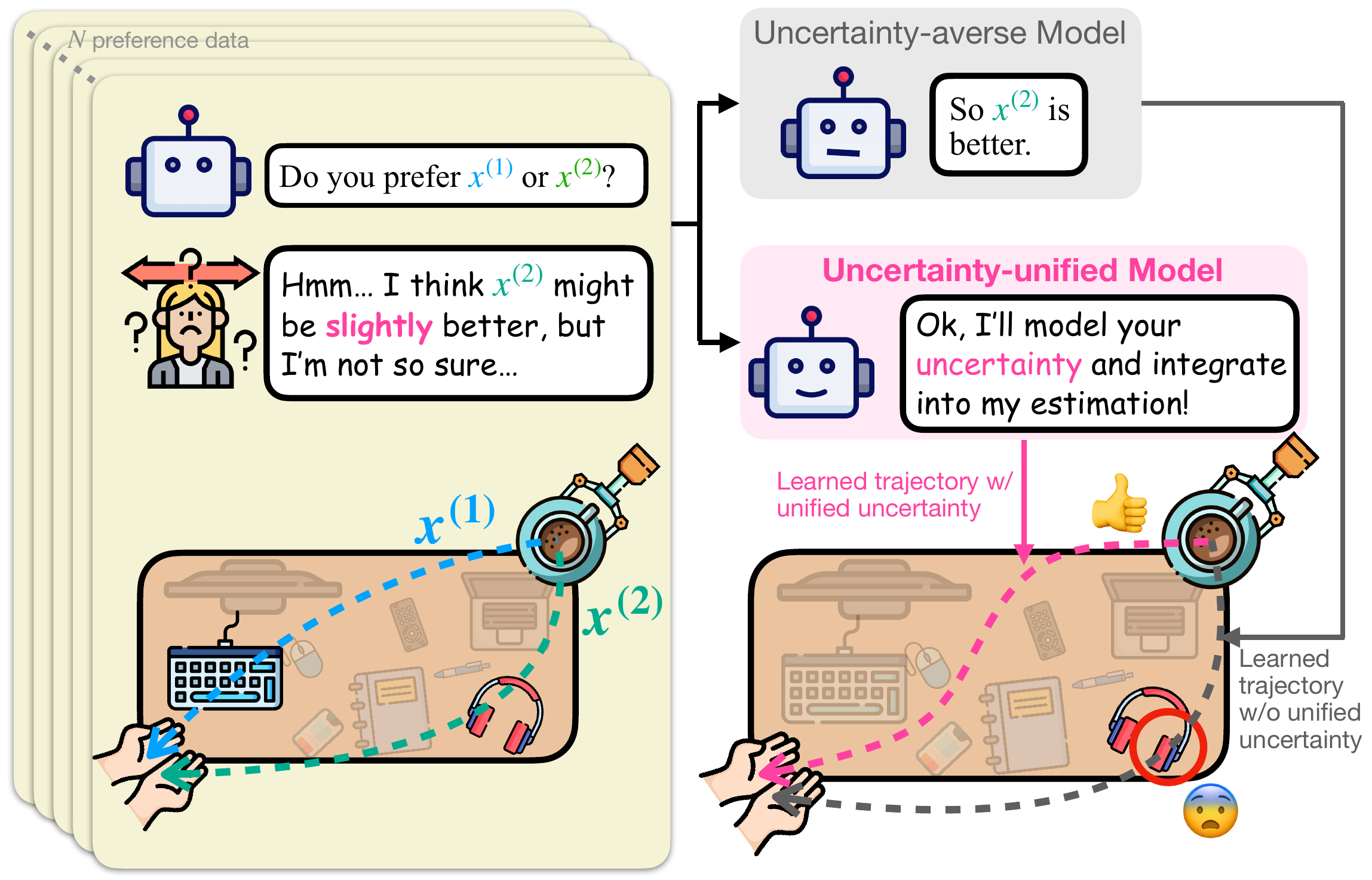}
    \vspace{-15pt}
    \caption{\textbf{Intuition on uncertainty unification for preference learning.} Imagine a robot inferring Alice's (a human user) ideal trajectory for passing a cup of coffee above a table using preference learning. In one sample pair, trajectory $x^{(1)}$ poses a risk of spilling coffee on the keyboard, while trajectory $x^{(2)}$ risks spilling it on the headphones. The keyboard and headphones are both valuable to Alice, so she responds that she weakly prefers $x^{(2)}$ with hesitation, reflecting her uncertainty in the decision. An uncertainty-averse model ignores this nuance, potentially learning a suboptimal and undesirable trajectory (e.g., still passing above the headphones). In contrast, an uncertainty-unified model incorporates Alice’s expressed uncertainty into its uncertainty-aware framework, enabling it to learn an ideal trajectory that aligns with her true preferences.}
    \label{fig:teaser}
    \vspace{-20pt}
\end{figure}

To date, research on robot and human uncertainties has largely progressed in isolation. This misalignment between two uncertainties can lead to inefficiencies, increased cognitive load, and even failures in critical decision-making scenarios like physical assistance and collaborations \cite{Medina2015SynthesizingAH, Yuan2022InSB}. Integrating these perspectives is vital for improving system adaptability, and ensuring robust, human-centric AI design. As this challenge is both technically complex and essential for advancing trustworthy human-robot partnerships, we argue a unified framework is necessary to tackle the misalignment. Thus, we define \textbf{uncertainty unification} in HRI as integrating human uncertainty into robotic uncertainty-aware algorithms, enabling outcomes informed by both human and robot uncertainties. To explore the potential of uncertainty unification, we focus on preference learning, a key HRI domain where robots learn from human feedback through comparative judgments. As suggested by \citet{Laidlaw2021UncertainDF}, this domain is well-suited for our study due to its rich uncertainty dynamics. Human uncertainty can be captured via self-reported confidence levels, while Gaussian Processes (GPs) are used to model robot's estimation of human preferences, with robot uncertainty quantified by the GP variances. Compared to neural network-based uncertainty estimations, GP variances offer advantages in interpretability, reliability, and sample efficiency \cite{Gawlikowski2021ASO, Zeng2024UncertaintyIF}. Fig.~\ref{fig:teaser} offers an intuitive example on the motivation of uncertainty unification in preference learning.

To unify human uncertainty with robotic Gaussian uncertainty, we propose uncertainty-unified preference learning (UUPL), which focuses on leveraging human confidence levels to refine the GP mean and covariance estimations. To this end, we design a human preference uncertainty model that captures varying levels of user's uncertainty, which is then integrated into the Laplace approximation algorithm to enhance GP mean estimation. Recognizing the limitations of the current GP predictive variance in preference learning (as further elaborated in Fig.~\ref{fig:uncertainty_exp}), we propose a human uncertainty-weighted Gaussian Mixture Model (GMM) to provide a more interpretable and adaptive GP variance estimation. This refined predictive variance not only achieves synergy with the acquisition function, but also has practical benefits in real-world applications. Additionally, we introduce an uncertainty calibration process to align uncertainty representations across diverse users, ensuring consistency and reliability in the model performance. Comprehensive simulations and real-world user studies, evaluated through accuracy metrics and Likert scale ratings, demonstrate the efficacy of UUPL. An ablation study further highlights the contributions of each uncertainty-unified component to overall system performance.

To sum up, the main contributions of our paper include:
\begin{itemize}
    \item We propose uncertainty-unified preference learning (UUPL) framework, which includes human preference uncertainty modeling and the integration into Laplace posterior mean estimation, uncertainty-weighted Gaussian Mixture Model for GP predictive variance scaling, and user-specific uncertainty calibration.
    \item We conduct comprehensive evaluations against three baselines across three simulation tasks, demonstrating significantly higher prediction accuracy. Additionally, we conduct three user studies, providing insights into the calibration process and illustrating practical applications of unified uncertainty. An ablation study further analyzes the contributions of individual components in UUPL.
\end{itemize}


\begin{figure*}[!t]
    \centering
    \includegraphics[width=0.99\textwidth]{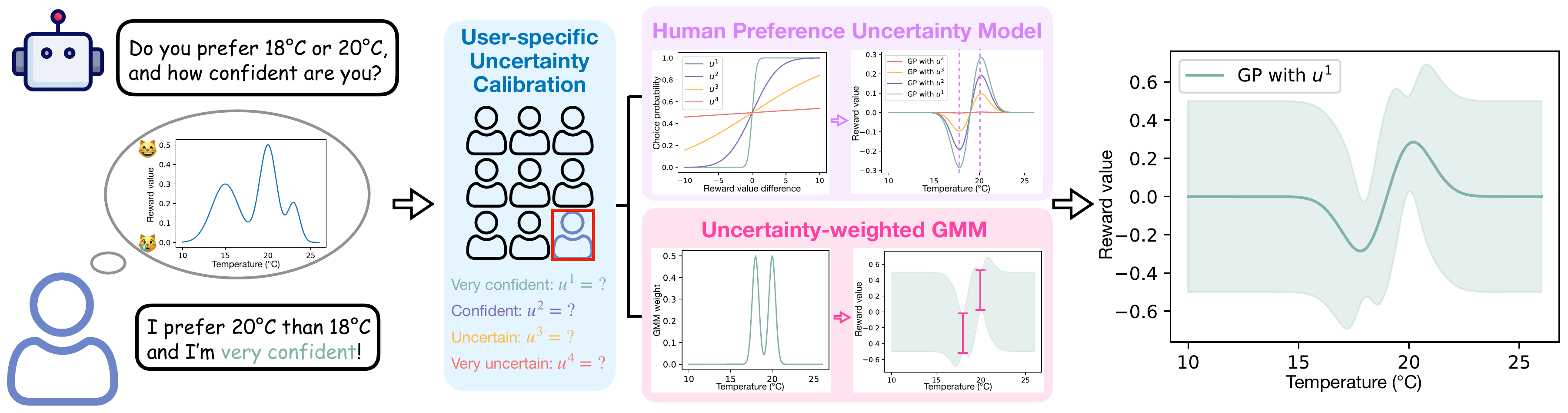} 
    \caption{\textbf{Overview of UUPL.} Imagine a robot inferring a user's preferred room temperature. For each query, we collect the user's preference with the associated uncertainty level. To begin, a calibration process (blue box) interprets the user's definitions of ``confident'' and ``uncertain'', ensuring these subjective assessments are accurately quantified with uncertainty factors $u$. Then, we construct the human preference uncertainty model as a probit model using Gaussian CDF, with the calibrated $u$ as the standard deviation (left part of purple box). This model improves the GP mean estimation accuracy (right part of purple box). Additionally, we introduce a weighted GMM (left part of red box) to adaptively scale the GP predictive variance (right part of red box) based on the human uncertainty level, enhancing its interpretability. Through this approach, UUPL effectively integrates human uncertainty into both the GP mean and variance, achieving comprehensive uncertainty unification, and thus provides a more accurate, interpretable, and user-aligned learning result (rightmost picture).}
    \vspace{-10pt}
    \label{fig:main} 
\end{figure*}

\section{Related Work}
\label{sec:rw}
This section first describes uncertainty quantification, followed by past work on modeling robot and human uncertainties in HRI, and finally discusses preference learning and its connection to uncertainty.

\subsection{Uncertainty Quantification and Modeling}
Uncertainty can broadly be categorized into two types: Aleatoric Uncertainty (AU) and Epistemic Uncertainty (EU) \cite{Senanayake2024TheRO}. AU, often referred to as stochastic or statistical uncertainty, represents inherent randomness in the system and cannot be mitigated through additional data or experiments. In contrast, EU, or systematic uncertainty, reflects a lack of knowledge about the system or environment and can potentially be reduced through further data collection.
From the robot's perspective, AU is generally quantifiable using traditional probabilistic methods like Monte Carlo simulations, and EU is typically estimated in Bayesian approaches. In Fig.~\ref{fig:teaser}, for example, consider the robot modeling the reward value for every possible trajectory. The uncertainties of trajectory $x^{(1)}$ or $x^{(2)}$'s reward values are dominated by AU because they have been observed, while the uncertainties of other unseen trajectories' reward values are dominated by EU. For human uncertainty in HRI, various methods have been explored to model AU. \citet{Laidlaw2021UncertainDF} employed inverse decision theory (IDT) to estimate human uncertainty, while \citet{Fisac2018ProbabilisticallySR} utilized a noisy-rationality model to predict human motion under uncertainty, recursively updating model confidence. Additionally, \citet{xu2024uncertaintyaware} incorporated human uncertainty into the Conditional Value-at-Risk (CVaR) framework to design an uncertainty-aware policy, and \citet{Holladay2016ActiveCB} proposed the Comparison Learning Algorithm for Uncertain Situations (CLAUS) based on the Luce-Shepard Choice Rule \cite{Logan2002AnIT}.
While significant progress has been made in AU modeling for both robots and humans, EU modeling for humans remains underexplored due to its inherently unmeasurable nature. Nevertheless, there is an emerging interest in how robot behaviors might provide humans with additional information about the environment, thereby indirectly reducing human EU. Importantly, most existing studies focus on either robot or human uncertainty in isolation, leaving a notable gap in achieving an integrated approach to uncertainty modeling in HRI. Although some work \cite{Fisac2018ProbabilisticallySR, Liu2015SafeEA, Hu2022ActiveUR} attempt to consider both human and robot uncertainties, they exclusively focus on the AU part, achieving only partial unification. Our work extends this by incorporating both aleatoric and epistemic components, thus providing a more general and broader uncertainty representation framework.

\subsection{Preference Learning}
Preference learning has become increasingly popular in robotics, though its definition can vary across domains. Following \citet{Frnkranz2010PreferenceLA}, we define preference learning as the task of learning a reward function from a set of comparison pairs with known preference relationships, formally defined in Section~\ref{sec:preferencelearning}. In the context of reinforcement learning (RL), preference learning can be seen as a generalization of inverse reinforcement learning (IRL), wherein the goal is to infer a reward function based on human feedback.
Several approaches have been proposed in preference learning. \citet{Sadigh2017ActivePL} represented the reward function as a weighted sum of features, updating the weights using Bayesian inference from human preferences. Nevertheless, this linear feature assumption limits the generalizability of the model. \citet{Chu2005PreferenceLW} employed Gaussian processes (GPs) to model reward functions, capturing nonlinearity. However, this approach lacked a strategy for generating next preference query. Later works, such as \citet{Bıyık2020AskingEQ} and \citet{Bıyık22024ActivePGP}, addressed this limitation by integrating predictive entropy to guide query selection, leveraging information gain principles as proposed by \citet{Houlsby2011BayesianAL}. Despite these advances, preference learning methods rarely account for human uncertainty. Some studies have explored weak preference modeling for human uncertainty. For example, \citet{Wilde2021LearningRF} used continuous-scale preference feedback to weight linear feature models, and \citet{Cao2021WeakHP} incorporated weak and equal preference options into deep neural network-based frameworks. However, these preference learning approaches focus solely on human uncertainty without considering robot uncertainty, leaving the challenge of integrating both sources of uncertainty unresolved. We bridge this gap by unifying uncertainties from both human and robot to achieve better preference learning results.

\section{Problem Formulation and Methods}
\label{sec:methods}
In this section, we begin by formulating the problem of preference learning and uncertainty unification. Then we describe how we model human uncertainty, providing an intuitive explanation of the approach. Finally, we introduce the GP framework for preference learning, and detail our uncertainty unification techniques with the calibration process.
\subsection{Problem Formulation}
\subsubsection{Preference Learning}
\label{sec:preferencelearning}
Given: (1) A sample space $\mathcal{O}$ such that the instances $O_i \in \mathcal{O}$ are any comparable data type (e.g., trajectory, state, object, $\dots$), (2) a feature extraction function $\phi: \mathcal{O} \to \mathbb{R}^n$ transforming each instance to a feature $x_{i} = \phi(O_{i}), x_{i} \in \mathbb{R}^n$, and (3) a human internal reward function $R(O_{i}): \mathcal{O} \to \mathbb{R}$ assigning each instance a real value. The goal of preference learning is to learn a function $f(x_{i}): \mathbb{R}^n \to \mathbb{R}$ to approximate $R(O_{i})$ from $N$ pairs of comparison data $\mathbf{X} = \{(x_1^{(1)}, x_1^{(2)}), (x_2^{(1)}, x_2^{(2)}), \dots, (x_N^{(1)}, x_N^{(2)})\}$ and the corresponding human choices $\mathbf{C} = \{C_1, C_2, \dots, C_N\}$, where $C_i = \{x_i^{(1)} \succ x_i^{(2)}\} \text{ if } R(O_i^{(1)}) > R(O_i^{(2)}), \text{and vice versa}$. $\{x_i^{(1)} \succ x_i^{(2)}\}$ denotes feature $x_i^{(1)}$ is preferred over feature $x_i^{(2)}$, implying that the corresponding option $O_i^{(1)}$ is also preferred over option $O_i^{(2)}$. In later sections, we use $O$ and $R(O)$ for models from human perspective, $x$ and $f(x)$ for models from robot perspective.

\subsubsection{Uncertainty Unification}
In general, when an HRI task includes, either in an explicit way or a probabilistic representation, both human uncertainty $u^H$ and robot uncertainty $u^R$, we say an algorithm $f$ is \textit{uncertainty unified} if it utilized both $u^H$ and $u^R$, i.e. $f(X, u^H, u^R) = y$, where $X, y$ are the original model input and output for the HRI task. This formulation is fundamentally different from \textit{uncertainty-aware} algorithms expressed as $f(X, u^R) = y$, which takes only robot uncertainty without considering human uncertainty \cite{Kim2023BridgingAE, Nguyen2024UncertaintyAV}. The intuition behind uncertainty unification is inspired by human-human interactions, where individuals often infer each other's confidence or uncertainty through verbal and non-verbal cues, using this information to guide joint decisions~\cite{Hsieh2019ConfidenceIB, Pejsa2014NaturalCU}. Similarly, uncertainty unification in HRI enables algorithms to model and mimic such communication, integrating both human and robot uncertainties into decision-making processes.


\subsection{Human Preference Uncertainty Modeling}
\label{sec:uahpm}
To achieve uncertainty unification in preference learning, it is essential to model the human uncertainty behind their choices. Given a query with two options $\{O^{(1)}, O^{(2)}\}$, in addition to collecting human choice $C$, we also collect the \textbf{human uncertainty level} $l$ and get the corresponding \textbf{human uncertainty factor} $u^l$ regarding this choice. In this work, we categorize human uncertainty levels into four discrete values: $l \in \{1~ \text{(very confident)},$ $ 2~ \text{(confident)}$, $3~ \text{(uncertain)}$, $4~ \text{(very uncertain)}\}$. These levels reflect the perceived similarity between $O^{(1)}$ and $O^{(2)}$. For instance, ``very uncertain'' implies that the options are nearly indistinguishable, while ``very confident'' indicates a strong preference. The uncertainty factor $u^l$ maps human uncertainty level to robot model parameters, and is further introduced in Section~\ref{Sec:usuc}. To represent the probability of a human choosing $O^{(1)}$ with uncertainty level $l$, we introduce a \textit{reward uncertainty residual} $r \sim \mathcal{N}(0, (u^l)^2)$:

\begin{equation}
\begin{aligned}
    & P(O^{(1)} (\succ^{l}) O^{(2)} | R(O^{(1)}), R(O^{(2)})) \\
    = & P(r < R(O^{(1)})-R(O^{(2)})) \\
    = & \Phi \left( \frac{R(O^{(1)})-R(O^{(2)})}{u^l} \right)
\label{eq:human_choice_model}
\end{aligned}
\end{equation}
where $\Phi$ is the cumulative distribution function (CDF) of the standard normal distribution. The model is shown as the left part of the purple box in Fig.~\ref{fig:main}.

\begin{figure}
    \centering
    \includegraphics[width=\columnwidth]{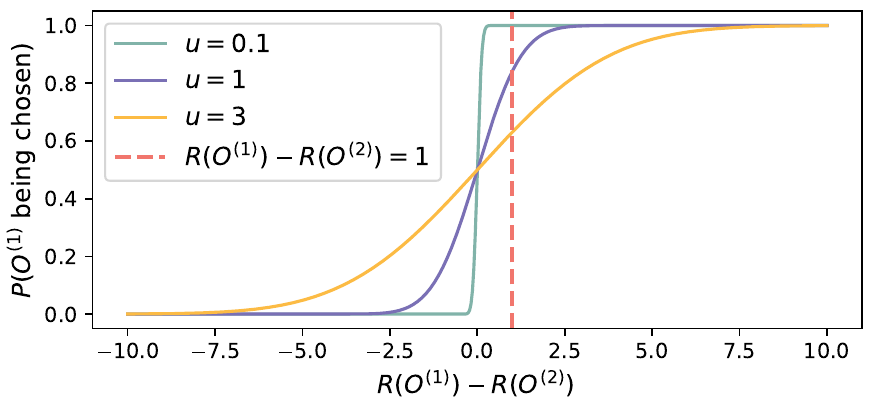}
    \vspace{-20pt}
    \caption{\textbf{Intuition behind our human preference uncertainty modeling.} The x-axis represents the reward residual $R(O^{(1)}) - R(O^{(2)})$, while the y-axis indicates the probability of selecting $O^{(1)}$. For a given query (marked by the red dashed line), confident human choices (low uncertainty level $l$ / small $u$) correspond to high probabilities, represented by the intersection between the red dashed line and the green CDF with $u = 0.1$. Conversely, uncertain human choices (high $l$ / large $u$) lower the modeled probability, as shown by the intersection between the red dashed line and the yellow CDF with $u = 3$.}
    \vspace{-15pt}
    \label{fig:u-Phi}
\end{figure}

Fig.~\ref{fig:u-Phi} further illustrates the relationship between the reward residual $R(O^{(1)})-R(O^{(2)})$ and the choice probability $P(O^{(1)} \text{being chosen})$ with multiple human uncertainty $u$ (modeled as the Gaussian variance $\sigma$). The more the human is uncertain, the less likely $O^{(1)}$ is chosen, as explained in the caption of Fig.~\ref{fig:u-Phi}.

\subsection{Gaussian Process for UUPL}
GPs are employed to model the human reward function, with their predictive variance serving as an indicator of uncertainty. In the context of preference learning, GPs estimate the uncertainty of the reward function across the entire feature space, and can be further integrated with the human preference uncertainty model described above. We develop a generalized framework for preference learning, built upon the Gaussian processes methodology introduced by \citet{Chu2005PreferenceLW} and \citet{Bıyık22024ActivePGP}, with these prior works emerging as special cases (i.e., zero human uncertainty) within our broader approach. In this section, we briefly introduce GPs, during which we emphasize how the human uncertainty is integrated into the framework to achieve unification. For more details on how GPs work in general, please refer to \cite{Rasmussen2005GaussianPF}.

\subsubsection{Kernel}
In a GP, the kernel $k(x_i^{(1)}, x_i^{(2)})$ is a positive semi-definite function that defines the covariance between any two feature points. In this work, we choose the most common radial basis function (RBF) kernel which is defined as
\begin{equation}
    k(x_i^{(1)}, x_i^{(2)}) = e^{-\gamma \|x_i^{(1)} - x_i^{(2)}\|^2}
\end{equation}
where $\gamma$ is a hyperparameter controlling kernel's smoothness. 

\subsubsection{Prior}
\label{Sec:prior}
Given $\mathbf{X} = \{(x_1^{(1)}, x_1^{(2)}), \dots, (x_N^{(1)}, x_N^{(2)})\}$, the corresponding predicted reward values are denoted by $\mathbf{f} = [f(x_1^{(1)}), f(x_1^{(2)}), \dots, f(x_N^{(1)}), f(x_N^{(2)})]^T$. Assuming a zero mean for $\mathbf{f}$, the prior is fully specified by the covariance matrix $\mathbf{K}$, which is a $2N \times 2N$ matrix with the $(ij)^{\text{th}}$ element be the kernel $k\left(x_{\lceil i/2 \rceil}^{(2-(i \text{ mod } 2))}, x_{\lceil j/2 \rceil}^{(2-(j \text{ mod } 2))}\right)$. The prior can then be expressed as a multivariate Gaussian:
\begin{equation}
\label{eq:f}
    P(\mathbf{f}) = \frac{1}{(2\pi)^{\frac{n}{2}}|\mathbf{K}|^{\frac{1}{2}}} e^{-\frac{\mathbf{f}^T \mathbf{K}^{-1} \mathbf{f}}{2}}
\end{equation}

\subsubsection{Likelihood}
The likelihood function takes the form of human preference uncertainty model as introduced in Section~\ref{sec:uahpm}. To express it from the robot's GP perspective, the likelihood for the $i^{\text{th}}$ comparison can be rewritten as
\begin{equation}
\begin{aligned}
    & P(x_i^{(1)} (\succ^{l_i}) x_i^{(2)} | f(x_i^{(1)}), f(x_i^{(2)})) \\
    = & \Phi \left( \frac{f(x_i^{(1)})-f(x_i^{(2)})}{u^{l_i}} \right)
\end{aligned}
\end{equation}

Furthermore, given a dataset $\mathbf{D} = \{(x_1^{(1)} (\succ^{l_1}) x_1^{(2)}), \dots,$ $ (x_N^{(1)} (\succ^{l_N}) x_N^{(2)})\}$, the likelihood for the dataset is the joint probability of observing each choice with reward $\mathbf{f}$, which can be written as the product of individual likelihood
\begin{equation}
\label{eq:df}
    P(\mathbf{D}|\mathbf{f}) = \prod_{i=1}^N \Phi \left ( \frac{f(x_i^{(1)})-f(x_i^{(2)})}{u^{l_i}}\right )
\end{equation}

\subsubsection{Posterior}
Based on Bayes' theorem, the posterior can be represented as 
\begin{equation}
    P(\mathbf{f} | \mathbf{D}) = \frac{P(\mathbf{f}) P(\mathbf{D} | \mathbf{f})}{P(\mathbf{D})}
\label{equ:posterior}
\end{equation}

Since a closed-form solution for this posterior is not available due to the existence of Gaussian CDFs in the likelihood~\cite{jensen2011pairwise}, we employ Laplace approximation: the mean of the new Gaussian refers to the maximum a posteriori estimate, and the variance refers to the inverse Hessian of the negative log-likelihood with respect to reward $f$. Next, we focus on introducing the integration of human uncertainty into Laplace approximation framework.

\subsubsection{Posterior Mean Approximation} Let the approximated GP mean be $\mathbf{f}_{\text{Lap}}$. Then, combining with Eq.~\ref{equ:posterior}, we have
\begin{equation}
\label{eq:pma}
\begin{aligned}
    \mathbf{f}_{\text{Lap}} & = \argmax_{\mathbf{f}} P(\mathbf{f} | \mathbf{D})\\
    & = \argmax_{\mathbf{f}} \sum^N_{i=1}\ln{\Phi \left ( \frac{f(x_i^{(1)}) - f(x_i^{(2)})}{u^{l_i}} \right )} - \frac{1}{2} \mathbf{f}^T \mathbf{K}^{-1} \mathbf{f}
\end{aligned}
\end{equation}

For more detailed derivation please refer to Appendix~\ref{apdx:1}. In Eq.~\ref{eq:pma}, the first part is a summation of $N$ monotonically increasing functions over differences of $\mathbf{f}$, and the second term is a quadratic penalization over $\mathbf{f}$. For simplicity, we set $N=1$, which means only one pair of preference data is collected. Then $\mathbf{f} = [f(x_1^{(1)}), f(x_1^{(2)})]^T \in \mathbb{R}^2, 
\mathbf{K} = \left[\begin{array}{cc} k(x_1^{(1)}, x_1^{(1)}) & k(x_1^{(1)}, x_1^{(2)}) \\ k(x_1^{(2)}, x_1^{(1)}) & k(x_1^{(2)}, x_1^{(2)}) \end{array}\right]
\in \mathbb{R}^{2 \times 2}$. 
Let $S(\mathbf{f}) = \ln{\Phi \left ( \frac{f(x_1^{(1)}) - f(x_1^{(2)})}{u^{l_1}} \right )} - \frac{1}{2} \mathbf{f}^T \mathbf{K}^{-1} \mathbf{f}$.

\begin{figure}[!t]
    \centering
    \includegraphics[width=\columnwidth]{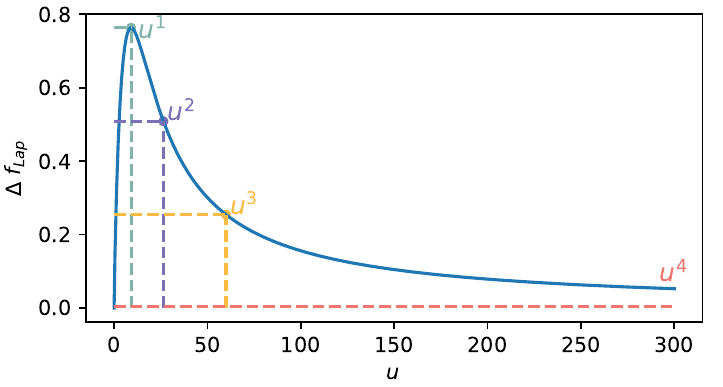}
    \vspace{-20pt}
    \caption{\textbf{Relationship between Laplace posterior mean difference $\boldsymbol{\Delta f_{\text{Lap}}}$ and human uncertainty $\boldsymbol{u}$.} $u^1, u^2, u^3, u^4$ are determined by varying the posterior mean difference proportionally, ensuring the model taking human uncertainty and producing more accurate estimated posterior mean.}
    \label{fig:uncertainty_level}
    \vspace{-10pt}
\end{figure}

Intuitively, the human uncertainty level of a choice reflects the difference of their internal reward function values of the query: greater uncertainty corresponds to a smaller reward difference, meaning the rewards of the two choices should be closer. To achieve this in Eq.~\ref{eq:pma}, we need to find a set of $u^l=\{u^1, u^2, u^3, u^4\}$, such that for each query $(x_i^{(1)}, x_i^{(2)})$, a larger $u^{l_i}$ results in a smaller $\Delta f_{\text{Lap}} = |{f}_{\text{Lap}}(x_1^{(1)}) - {f}_{\text{Lap}}(x_1^{(2)})|$, which requires us to find the relationship between $u$ and the posterior mean difference $\Delta f_{\text{Lap}}$. 

Since solving $\frac{d}{d\mathbf{f}}S(\mathbf{f}) = 0$ directly is intractable due to the Gaussian CDF $\Phi$, we approach it analytically by determining $\Delta f_{\text{Lap}}$ against various $u$. Denote $d$ as the monotonically decreasing part of $\Delta f_{\text{Lap}}$ (i.e., where $u \ge 10$), then we design $u^l = \{u^1, u^2, u^3, u^4\}$ to be the four percentage points of $d(u)$ as shown in Fig.~\ref{fig:uncertainty_level}. When human uncertainty level $l=1~\text{(very confident)}$, $u^1$ takes $d^{-1}(d_{\text{max}})$. When human uncertainty level $l=2~\text{(confident)}$, $u^2$ takes $d^{-1}(\frac{2}{3}d_{\text{max}})$. When human uncertainty level $l=3~\text{(uncertain)}$, $u^3$ takes $d^{-1}(\frac{1}{3}d_{\text{max}})$. Lastly, when human uncertainty level $l=4~\text{(very uncertain)}$, $u^4$ takes $d^{-1}(1e-3)$. This mapping ensures that as human uncertainty $l$ increases, the posterior mean difference between rewards $\Delta f_{\text{Lap}}$ decreases proportionally, aligning with the intuition of uncertain preferences. The resulting posterior GP mean with different human uncertainty levels is shown as the right part of the purple box in Fig.~\ref{fig:main}.

\textbf{User-specific Uncertainty Calibration}\label{Sec:usuc}: Given individual differences in defining ``confident'' or ``uncertain'', a fixed mapping from $l$ to $u^l$ may be inadequate. Thus, we propose a calibration process to tailor $u^l$ values to individual users, by asking each user to describe a given reward function and customize their uncertainty factors from their answers. Specifically, we design a calibration function $f_{\text{calib}}$ and perform a data collection process of $M$ iterations: At iteration $i$, a feature pair $(x_i^{(1)}, x_i^{(2)})$ is sampled and the user is told to assume the function values $(f_{\text{calib}}(x_i^{(1)}), f_{\text{calib}}(x_i^{(2)}))$ reflect their internal reward values, i.e., $f_{\text{calib}}(x_i^{(k)}) = R(x_i^{(k)}), k \in \{1,2\}$. Then the user specifies their uncertainty level $l$, and the corresponding value difference $\Delta f_{\text{calib}} = |f_{\text{calib}}(x_i^{(1)}) - f_{\text{calib}}(x_i^{(2)})|$ is saved together with their chosen uncertainty level $l_i \in \{1,2,3,4\}$. After collecting $M$ pairs of data, the mean value difference for each uncertainty level $\overline{\Delta f}_{\text{calib}}^l$ and the corresponding quantiles $q^l = \frac{\overline{\Delta f}_{\text{calib}}^l - \text{min}(f_{\text{calib}})}{\text{max}(f_{\text{calib}}) - \text{min}(f_{\text{calib}})}, l \in \{1,2,3,4\}$ are calculated. We then get $\Delta f_{\text{Lap}}^l$ and further more the corresponding $u^l$ with the same quantile on the function as in Fig.~\ref{fig:uncertainty_level}. The pseudo-code for the calibration process is provided as Algorithm~\ref{code:1}.


\subsubsection{Posterior Covariance Approximation}
To approximate the GP covariance, we take the negative inverse Hessian of the logarithm of the un-normalized posterior with respect to $\mathbf{f}$~\cite{Bıyık22024ActivePGP, Rasmussen2005GaussianPF, jensen2011pairwise}. Let the approximated covariance be $\mathbf{K}_{\text{Lap}}$, then
\begin{equation}
\begin{aligned}
    \mathbf{K}_{\text{Lap}} &= -(\nabla \nabla(\ln P(\mathbf{D}|\mathbf{f}) + \ln P(\mathbf{f})))^{-1} \\
    &= (\mathbf{W} + \mathbf{K}^{-1})^{-1}
\end{aligned}
\end{equation}
where $\mathbf{W}$ is the negative Hessian of the log-likelihood, and $\mathbf{K}$ is the $2N \times 2N$ covariance matrix of $N$ preference pairs.

\newcommand{\nosemic}{\renewcommand{\@endalgocfline}{\relax}}
\newcommand{\dosemic}{\renewcommand{\@endalgocfline}{\algocf@endline}}
\newcommand{\pushline}{\Indp}
\newcommand{\popline}{\Indm\dosemic}
\let\oldnl\nl
\newcommand{\nonl}{\renewcommand{\nl}{\let\nl\oldnl}}
\newcommand\mycommfont[1]{\footnotesize\ttfamily\textcolor{blue}{#1}}
\SetCommentSty{mycommfont}
\SetKwInOut{Input}{Input}
\SetKwInOut{Output}{Output\,}
\SetKwInOut{Parameters}{Parameters}
\SetKwProg{Tree}{Tree}{}{EndTree}
\setlength{\textfloatsep}{0pt}
\begin{algorithm}[!t]
\label{code:1}
    \small{
    \caption{User-specific Uncertainty Calibration}
    \Input{
    $f_\text{calib}$; $\Delta f_{\text{Lap}}(u)$ as in Fig~\ref{fig:uncertainty_level}; $\mathbf{X}_{\text{calib}}$ with uniformly generated $(\mathbf{x}^{(1)}, \mathbf{x}^{(2)})$
    }
    \Output{$u^1, u^2, u^3, u^4$
    }
    \textbf{Init} $\text{list}_{u^1} \gets$ [], $\text{list}_{u^2} \gets$ [], $\text{list}_{u^3} \gets$ [], $\text{list}_{u^4} \gets$ []

    \For{$(x_i^{(1)}, x_i^{(2)})$ \textbf{\textup{in}} $\mathbf{X}_{\mathrm{calib}}$}{
    Collect user uncertainty level $l_i$ \\
    list$_{u^{l_i}}$.add$(|f_{\text{calib}}(x_i^{(1)}) - f_{\text{calib}}(x_i^{(2)})|)$
    }
    \For{$l$ \textbf{\textup{in}} $\{1,2,3,4\}$}{
    \tcc{Get mean value}
    $\overline{\Delta f}_{\text{calib}}^l \gets $\text{avg}$(\text{list}_{u^l})$ \\
    \tcc{Get quantile}
    $q^l \gets \frac{\overline{\Delta f}_{\text{calib}}^l - \text{min}(f_{\text{calib}})}{\text{max}(f_{\text{calib}}) - \text{min}(f_{\text{calib}})}$ \\
    \tcc{Get $u^l$ with the same quantile}
    $d \gets$ the monotonically decreasing part of $\Delta f_{\text{Lap}}$ \\
    $u^l \gets d^{-1} (q^l \cdot d_{\text{max}})$
    }
    }
\end{algorithm}

\subsubsection{Uncertainty-unified Prediction}
\label{Sec:prediction}
One benefit of GP is that it provides a closed-form solution for prediction. Suppose we have a test preference pair $\mathbf{x}_t = (x_t^{(1)}, x_t^{(2)})$, let the predictive mean be $\boldsymbol{\mu}_t = [f(x_t^{(1)}), f(x_t^{(2)})]^T$ , and the covariance be $\boldsymbol{\Sigma}_t = \left[\begin{array}{cc} \Sigma_t^{(1)(1)} & \Sigma_t^{(1)(2)} \\ \Sigma_t^{(2)(1)} & \Sigma_t^{(2)(2)} \end{array}\right] $, then we have: 
\setlength{\arraycolsep}{0.9pt}
\begin{equation}
    \boldsymbol{\mu}_t = \mathbf{k}_{t}^T\mathbf{K}^{-1}\mathbf{f}_{\text{Lap}}
\label{eq:10}
\end{equation}
\begin{equation}
    \boldsymbol{\Sigma}_t = \mathbf{K}_t - \mathbf{k}_{t}^T (\mathbf{W} + \mathbf{K}^{-1})^{-1} \mathbf{k}_{t}
\label{eq:11}
\end{equation}
where $\mathbf{k}_{t} = \left[\begin{array}{cccc} k(x_t^{(1)}, x_1^{(1)}), & k(x_t^{(1)}, x_1^{(2)}), & \dots, & k(x_t^{(1)}, x_N^{(2)}) \\ k(x_t^{(2)}, x_1^{(1)}), & k(x_t^{(2)}, x_1^{(2)}), & \dots, & k(x_t^{(2)}, x_N^{(2)}) \end{array}\right]^T$ and 
\setlength{\arraycolsep}{5pt}
$\mathbf{K}_t = \left[\begin{array}{cc} k(x_t^{(1)}, x_t^{(1)}), & k(x_t^{(1)}, x_t^{(2)}) \\ k(x_t^{(2)}, x_t^{(1)}), & k(x_t^{(2)}, x_t^{(2)}) \end{array}\right]$.

By integrating $\mathbf{f}_{\text{Lap}}$, the predictive mean $\boldsymbol{\mu}_t$ in Eq.~\ref{eq:10} has already taken human uncertainty into account, which can generate more accurate reward values. While for the predictive covariance $\boldsymbol{\Sigma}_t$ in Eq.~\ref{eq:11} we introduce a Gaussian Mixture Model (GMM) based method to unify human uncertainty with robot uncertainty, as introduced below.

\textbf{Uncertainty-weighted Gaussian Mixture Model}
\label{uupc}
Under the uncertainty-unification setting, the GP covariance $\text{Cov}(f(x_t^{(1)}), f(x_t^{(2)}))$ should reflect both robot and human uncertainty for any test feature pair $(x^{(1)}_t, x^{(2)}_t)$. Robot uncertainty is expressed by the predictive covariance $\boldsymbol{\Sigma}_t$ as Eq.~\ref{eq:11}, and we design human uncertainty to be a GMM $\mathcal{G}$ built upon all the $N$ (query, uncertainty) pairs $\{((x_1^{(1)}, x_1^{(2)}), u^{l_1}), \dots,((x_N^{(1)}, x_N^{(2)}), u^{l_N})\}$.

\begin{equation}
\begin{aligned}
    \mathcal{G}(x^{(1)}_t) = 1 + \sum_{i=1}^{N} \sum_{k=1}^{2} w(u^{l_i}) \mathcal{N}(x^{(1)}_t; x_i^{(k)}, \sigma^2) \\
    \mathcal{G}(x^{(2)}_t) = 1 + \sum_{i=1}^{N} \sum_{k=1}^{2} w(u^{l_i}) \mathcal{N}(x^{(2)}_t; x_i^{(k)}, \sigma^2)
\end{aligned}
\end{equation}
where $w$ is a bijection mapping $\{u^1, u^2, u^3, u^4\}$ to $\{w^1, w^2, w^3, w^4\}$ representing decreasing weights for increasing uncertainty level\footnote{$w^{1,2,3,4}$ are hyperparameters determined empirically such that $w^1>w^2>w^3>w^4>0$ and $\mathcal{G}(x^{(m)}_t) \ge 1, m \in \{1,2\}$.}. Then, we define the uncertainty-unified predictive covariance matrix $\boldsymbol{\Sigma}_t^\prime$ as:
\begin{equation}
\begin{aligned}
    \boldsymbol{\Sigma}_t^\prime = \begin{bmatrix} \mathcal{G}(x_t^{(1)}) & 0 \\ 0 & \mathcal{G}(x_t^{(2)}) \end{bmatrix}^{-1}
    \boldsymbol{\Sigma}_t
    \begin{bmatrix} \mathcal{G}(x_t^{(1)}) & 0 \\ 0 & \mathcal{G}(x_t^{(2)}) \end{bmatrix}^{-1}
\end{aligned}
\end{equation}
or equivalently
\begin{equation}
\label{eq:var}
\begin{aligned}
    \text{Var}^\prime(f(x_t^{(m)})) = \mathcal{G}(x_t^{(m)})^{-2} \boldsymbol{\Sigma}_{t}^{(m)(m)}, m \in \{1,2\}\\
    \text{Cov}^\prime(f(x_t^{(1)}), f(x_t^{(2)})) = \mathcal{G}(x_t^{(1)})^{-1} \mathcal{G}(x_t^{(2)})^{-1} \boldsymbol{\Sigma}_{t}^{(1)(2)}
\end{aligned}
\end{equation}

\begin{figure*}[!t]
    \centering
    \includegraphics[width=\textwidth]{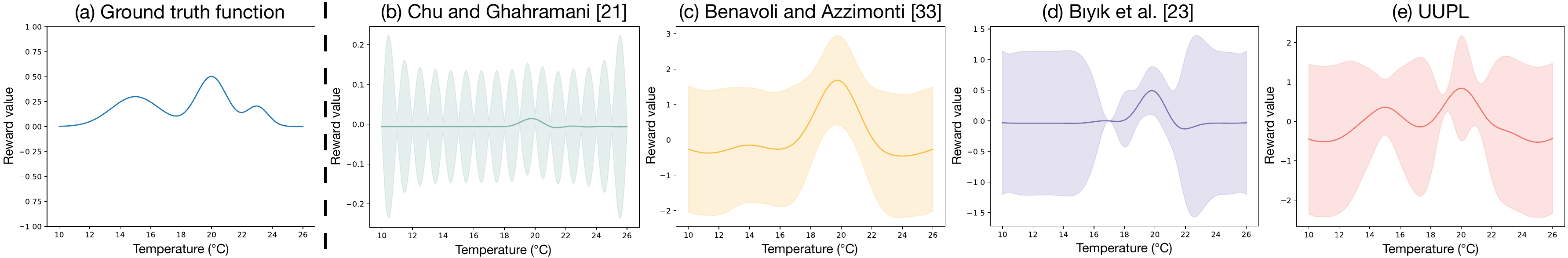} 
    \vspace{-15pt}
    \caption{\textbf{GP variance visualizations.} The ground truth function and the learned GPs (mean $\pm$ $1.96\times$std) of three baseline methods and UUPL are provided. The data comes from comparisons of 19°C with all other integer temperatures, and the results showcase the rationality of our learned variance.}
    \label{fig:uncertainty_exp} 
    \vspace{-10pt}
\end{figure*}

An example of uncertainty-weighted GMM $\mathcal{G}$ and the scaled predictive variance $\text{Var}^\prime$ is shown in the red box in Fig.~\ref{fig:main}. Intuitively, all preference pairs and the corresponding human uncertainty levels impact the variance of $f(x_t^{(m)}), m \in \{1,2\}$: If many query points are observed near $x_t^{(m)}$ (i.e., large value for $\mathcal{N}(x_t^{(m)}; x_i^{(k)}, \sigma^2)$), then $\text{Var}(f(x_t^{(m)}))$ should be scaled down; If the human has relatively low uncertainty level $l$ (i.e., high confidence) for the preferences, then $x_t^{(m)}$ should also reflect a relatively low uncertainty from the robot's perspective (i.e., large value for $w(u^{l_i})$). Conversely, if the observed points are all far away from $x_t^{(m)}$, or the human demonstrates high uncertainty (i.e., low confidence), then the uncertainty of the test point should also be high. We claim that this unified design provides a rational variance estimation that benefits both theoretical acquisition functions (described in the next section) and practical applications in robotic experiments (Section~\ref{sec:us}). Variance analyses in Section~\ref{sec:tcs} and Fig.~\ref{fig:uncertainty_exp} further validate the effectiveness of uncertainty-weighted GMM.

\subsubsection{Acquisition Function}
The acquisition function generates a new data pair to query the user preference and uncertainty. \citet{Bıyık22024ActivePGP} proposed an acquisition function that maximizes the query information gain and at the same time minimizes human's burden of answering the question \cite{Bıyık2020AskingEQ}. Here, we improve this design with our uncertainty-unified variance $\text{Var}^\prime$ and covariance $\text{Cov}^\prime$. Let the next query be ${x}_{*} = ({x}_{*}^{(1)}, {x}_{*}^{(2)})$ and $H$ be the information entropy, then:
\begin{equation}
\begin{aligned}
    x_{*} &= \argmax_{x^{(1)}, x^{(2)}} I(f; C|Q,\mathbf{D}) \\
    &= \argmax_{x^{(1)}, x^{(2)}} (H(C|Q, \mathbf{D}) - E_{f \sim P(f|\mathbf{D})}[H(C|Q,f)])
\end{aligned}
\end{equation}
which can be further written as
\begin{equation}
\begin{aligned}
    h\left( \Phi \left(\frac{\mu^{(1)} - \mu^{(2)}}{\sqrt{{(u^l)}^2 + g(\mathbf{x}^{(1)}, \mathbf{x}^{(2)})}} \right)\right) - m(\mathbf{x})
\end{aligned}
\end{equation}
where $h$ is the binary entropy function, $g$ and $m$ can be written as
\begin{equation}
\begin{aligned}
        g(\mathbf{x}^{(1)}, \mathbf{x}^{(2)}) = ~& \text{Var}^\prime(f(\mathbf{x}^{(1)})) + \text{Var}^\prime(f(\mathbf{x}^{(2)})) \\ 
    &- 2 \text{Cov}^\prime(f(\mathbf{x}^{(1)}), f(\mathbf{x}^{(2)}))
\end{aligned}
\end{equation}
\begin{equation}
\begin{aligned}
    m(\mathbf{x}) = \frac{\sqrt{\pi \ln(2) {(u^l)}^2} \exp{\left( -\frac{(\mu^{(1)} - \mu^{(2)})^2}{\pi \ln(2) {(u^l)}^2 + 2g(\mathbf{x}^{(1)},\mathbf{x}^{(2)})} \right )}}{\sqrt{\pi \ln(2) {(u^l)}^2 + 2g(\mathbf{x}^{(1)},\mathbf{x}^{(2)})}}
\end{aligned}
\end{equation}
where $\text{Var}^\prime$ and $\text{Cov}^\prime$ are defined in Eq.~\ref{eq:var}. For a more detailed derivation of such result, please refer to \citet{Bıyık22024ActivePGP}. 

It is important to emphasize that this acquisition function is designed to select the next query based on two key criteria: similar reward values and high variances. In traditional binary preference settings, users may struggle to make selections when presented with two options with similar rewards (e.g., $x^{(1)}$ and $x^{(2)}$ in Fig.~\ref{fig:teaser}). By incorporating uncertainty, our approach improves the user experience by allowing users to express their confidence levels alongside their choices. This claim will be supported later in user study sections.

Additionally, under the uncertainty-unification framework, the GP variance is further enriched by integrating human uncertainty. This enhances the representativeness of the selected query by maximizing information gain, taking into account the joint uncertainties of both humans and robots. For example, when faced with two queries observed the same number of times (i.e., identical robot uncertainty), our model prioritizes the one with higher human uncertainty, as it exhibits greater entropy. Intuitively, this means the robot is more likely to generate queries labeled as ``uncertain'' rather than ``confident'', thereby focusing on grounding the uncertain areas. As a result, the proposed method selects more informative queries, which we hypothesize will accelerate the convergence of the GP model. This claim is validated in the experiment section.


\section{Experiment}
The primary contribution of this work is demonstrating that uncertainty unification significantly enhances the performance of preference learning tasks. We conduct comprehensive experiments to validate the effectiveness of our method (UUPL) both quantitatively and qualitatively. First, we design three simulation experiments to illustrate the rationality of our uncertainty-unified framework, and analyze its accuracy compared to other baseline methods. Following, we present an ablation study which investigates the contributions of individual components. Finally, we present three user studies to demonstrate the efficacy of UUPL, especially our user-specific uncertainty calibration (Section~\ref{Sec:usuc}) in real-world settings, and analyze user feedback through Likert scale ratings. 

\subsection{Simulation Experiments}
\label{sec:sim}
For simulation experiments, we design ground truth functions to simulate human reward functions and decide choices $\mathbf{C}$ with uncertainty levels $\mathbf{l}$ based on Eq.~\ref{eq:human_choice_model}. The human uncertainty $u = \{u^1, u^2, u^3, u^4\}$ is determined proportionally according to Section~\ref{Sec:usuc}. Each experiment is repeated six times, with $N$ iterations per trial ($N=50$ for Simulations 1 and 2, $N=100$ for Simulation 3). In each iteration, $C$ is generated from the relative function values, $l$ is calculated from the function value differences, and finally, the GP is updated. We record the accuracies and their variances across the 6 trials. The goal is to compare the learned functions from different methods with the ground truth function.

\subsubsection{Metrics \& Baselines}
With a fine granularity, we discretize the learned GP mean and the ground truth function such that $F_{\text{pred}} = [\mu_{\text{pred}}(x_1), \mu_{\text{pred}}(x_2), \dots, \mu_{\text{pred}}(x_n)]$ and $F_{\text{gt}} = [f_{\text{gt}}(x_1), f_{\text{gt}}(x_2), \dots, f_{\text{gt}}(x_n)]$, with the corresponding means be $\overline{\mu}_{\text{pred}}$ and $\overline{f}_{\text{gt}}$. Then we use the sample correlation coefficient $r$ to be our accuracy metric for simulation experiments:
\begin{equation}
    r = \frac{\sum_{i=1}^n (\mu_{\text{pred}}(x_i) - \overline{\mu}_{\text{pred}})(f_{\text{gt}}(x_i) - \overline{f}_{\text{gt}})}{\sqrt{\sum_{i=1}^n (\mu_{\text{pred}}(x_i) - \overline{\mu}_{\text{pred}})^2 \sum_{i=1}^n (f_{\text{gt}}(x_i) - \overline{f}_{\text{gt}})^2}}
\end{equation}

The sample correlation coefficient is scale-invariant and focuses on the \textit{trend} of the data. This property is ideal for evaluating preference pairs, as the absolute alignment of $F_{\text{pred}}$ and $F_{\text{gt}}$ is less critical than their relative ordering and overall trajectory (i.e., ensuring that the functions evolve similarly).

For the baseline methods, we choose three other GP-based preference learning methods: Baseline 1 as \citet{Chu2005PreferenceLW}, Baseline 2 as \citet{benavoli2024tutorial}, and Baseline 3 as \citet{Bıyık22024ActivePGP}, each with different kernel designs, approximation methods, and acquisition functions. Note that none of these baselines are uncertainty unified.

\subsubsection{Simulation 1 -- Thermal Comfort}
\label{sec:tcs}
\begin{figure*}[!t]
    \centering
    \includegraphics[width=\textwidth]{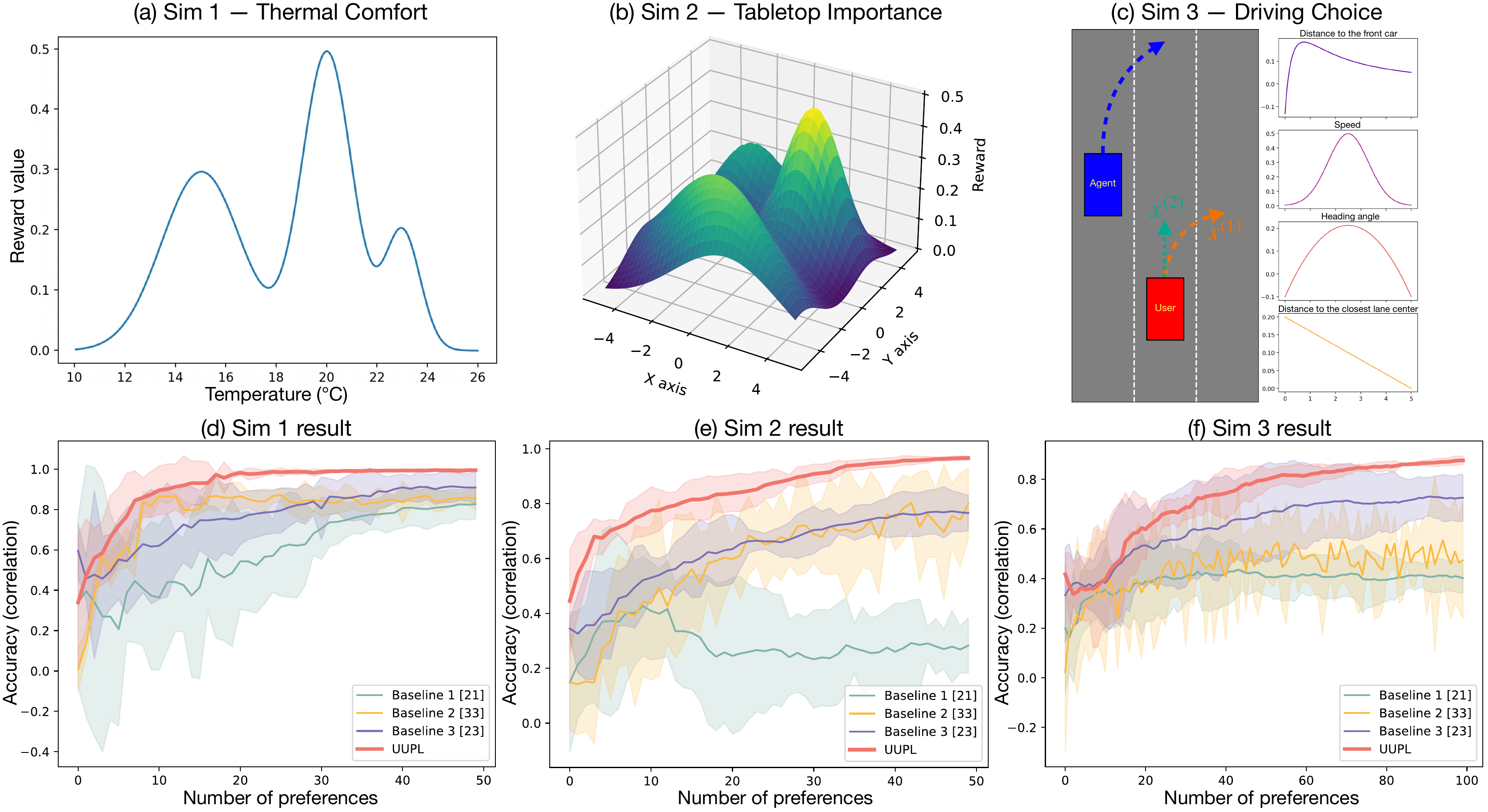}
    \vspace{-15pt}
    \caption{\textbf{Three simulation experiments and their results.} (a) The 1D thermal comfort function used in Simulation 1, with results shown in (d). (b) The 2D tabletop importance function for Simulation 2, with results plotted in (e). (c) Simulation 3, where the left panel depicts two possible trajectories for the red user car based on the blue agent car's action, and the right panel shows the functions of the four trajectory-related features. Results are presented in (f).}
    \label{fig:exps}
\end{figure*}

The first simulation experiment focuses on a conceptual scenario of learning a user’s preferred room temperature, inspired by \citet{benavoli2024tutorial}. The feature is a 1D scalar representing the temperature in Celsius: $x_{i} \in [10, 26]$, and the ground truth function $f(x_{i})$ is illustrated in Fig.~\ref{fig:exps}(a). This function reflects localized preferences for room temperature based on activities such as cooking, working, and sleeping.

We first demonstrate the rationality of our uncertainty-unified predictive variance (Section~\ref{uupc}) qualitatively, leveraging the intuitiveness of the 1D variance. The results are shown in Fig.~\ref{fig:uncertainty_exp}. Suppose in (a) the feature $x = 19$ (i.e., 19°C) has been observed 17 times, with comparisons made against uniformly selected features $x=10, 11, \dots, 26$. \citet{Chu2005PreferenceLW} assign zero variances to all the observed points, despite most of them only appeared once. \citet{benavoli2024tutorial} assign uniformly high variances across all points, ignoring the observation effects. \citet{Bıyık22024ActivePGP} initialize a reference point $f(17)$ with zero variance due to its kernel design, which does not contain insightful meaning and requires careful hyperparameter tuning. In contrast, UUPL is tuning-free and assigns the smallest variance to $f(19)$, effectively capturing the impact of observations on variance reduction and improving the variance interpretability. Moreover, our human preference uncertainty model allows the learned function to quickly identify two local maxima, whereas baseline methods only detect a single maximum. This demonstrates the representational power of our uncertainty-unified framework.

\begin{table*}[htbp]
\centering
\caption{\textnormal{\small{Accuracies $(\text{mean} \pm \text{std})$ of three baseline methods and UUPL on the three simulation experiments.}}}
\vspace{-5pt}
\begin{tabular}{lcccc} 
\toprule
& Baseline 1~\cite{Chu2005PreferenceLW} & Baseline 2~\cite{benavoli2024tutorial} & Baseline 3~\cite{Bıyık22024ActivePGP} & \textbf{UUPL} \\ \midrule
Sim 1 -- Thermal Comfort & $0.8260 \pm 0.0370$ & $0.8523 \pm 0.0281$ & $0.9087 \pm 0.0402$ & $\mathbf{0.9949} \bm{\pm} \mathbf{0.0024}$ \\
Sim 2 -- Tabletop Importance & $0.2830 \pm 0.0515$ & $0.7997 \pm 0.0655$ & $0.7659 \pm 0.0332$ & $\mathbf{0.9653} \boldsymbol{\pm} \mathbf{0.0039}$ \\
Sim 3 -- Driving choice & $0.4018 \pm 0.0293$ & $0.4733 \pm 0.1202$ & $0.7254 \pm 0.0479$ & $\mathbf{0.8764} \bm{\pm} \mathbf{0.0094}$ \\ \bottomrule
\end{tabular}
\vspace{-10pt}
\label{tab}
\end{table*}

\subsubsection{Simulation 2 -- Tabletop Importance}
This simulation explores a more realistic setting: a tabletop scenario similar to Fig.~\ref{fig:teaser}. Imagine a user sitting at a table cluttered with various objects, such as electronics, stationery, and snacks, while a home robot attempts to deliver a cup of coffee from the opposite side of the table. To successfully complete the task, the robot must learn the user's preferences to minimize the risk of spilling coffee onto critical objects. Directly assigning scores to all points on the tabletop is impractical for users, and general ``common sense'' rules (e.g., electronics are often considered to be more important than snacks) often fail to account for individual preferences, making preference learning with GP an appropriate choice.

The task, illustrated in Fig.~\ref{fig:exps}(b), models the tabletop as a 2D feature space where $x_i \in ([-5, 5] \times [-5, 5])$. Three hills of varying shapes and scales represent objects on the table, with the function value at each point indicating the importance at the corresponding location -- higher values correspond to higher importance. It is worth noting that, compared to directly comparing trajectory features as in Fig.~\ref{fig:teaser}, our design offers additional insights, such as the uncertainty in specific areas of the tabletop. This information can be further leveraged and will be demonstrated in the first robot experiment in Section~\ref{sec:robotexp}. Moreover, the desired trajectory can be computed using off-the-shelf path planning algorithms, seamlessly integrating the learned reward function into the robot's planning process.


\subsubsection{Simulation 3 -- Driving Choice}

The last simulation builds on scenarios described by \citet{Sadigh2016PlanningFA} and \citet{Bıyık22024ActivePGP}. As illustrated by Fig.~\ref{fig:exps}(c), an agent (blue) car and a user (red) car navigate a three-lane road. The agent car intends to switch from the left to the middle lane, prompting the user car to respond by slowing down, switching lanes, or adopting other behaviors. At each iteration, two possible trajectories (green and orange arrows) are presented. The features are designed to consider four important aspects: Distance to the front car, speed, heading angle, and distance to the closest lane center (all normalized to $[0,5]$), so $x_{i} \in ([0,5] \times [0,5] \times [0,5] \times [0,5])$. We design four complex functions for each feature as shown on the right part of Fig.~\ref{fig:exps}(c), and the ground truth reward function $f(x_i): \mathbb{R}^4 \to \mathbb{R}$ is the sum of four function values for each feature. Due to the experiment's complexity, the number of preference collection was increased to 100.


\subsubsection{Simulation Results Analysis}

\begin{figure*}[!t]
    \centering
    \includegraphics[width=\textwidth]{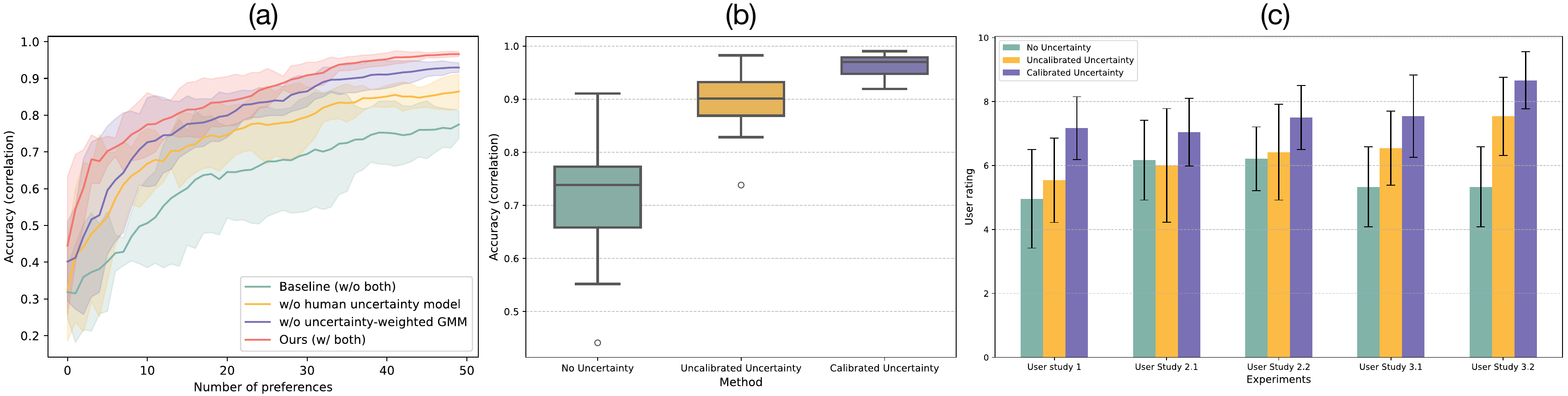}
    \vspace{-15pt}
    \caption{\textbf{(a) Ablation study results:} The red curve represents UUPL, the purple curve excludes the uncertainty-weighted GMM, the yellow curve omits the human uncertainty model, and the green curve is the baseline without modeling uncertainty. \textbf{(b) Calibration evaluation:} The ``No uncertainty'' method shows the lowest accuracy and highest variance, while our ``Calibrated uncertainty'' method achieves the highest accuracy and lowest variance. \textbf{(c) User study ratings:} Across all user studies, our ``Calibrated uncertainty'' method outperforms others by at least one rating scale.}
    \label{fig:real_results}
    \vspace{-10pt}
\end{figure*}

The mean and variance of the accuracy for all three simulations are presented in Fig.~\ref{fig:exps}(d)-(f) and Table~\ref{tab}. To provide a comprehensive evaluation, we analyze the results from three perspectives: \textbf{accuracy}, \textbf{efficiency}, and \textbf{stability}.
\begin{itemize}
    \item \textbf{Accuracy}: For all three simulations, UUPL demonstrates the highest final accuracy, achieving 0.9949, 0.9653, and 0.8764 for Simulation 1, 2, and 3, respectively. This superior performance is largely attributed to the human preference uncertainty model. By allowing users to express uncertainty levels and integrating this information into the Laplace approximation, UUPL refines the posterior mean estimation. As shown in the purple box in Fig.~\ref{fig:main}, this integration ensures the posterior mean is scaled more accurately based on the uncertainty levels, leading to improved accuracies across tasks.
    \item \textbf{Efficiency}: UUPL exhibits the fastest convergence rates. For Simulation 1, UUPL reaches an accuracy of 0.7 within just 5 iterations. Although Baseline 2 achieves a comparable performance of 8 iterations, its results deteriorate after reaching 0.8 accuracy. Baseline 3 and 1 require 13 and 30 iterations respectively to reach 0.7 accuracy. In Simulation 2, UUPL attains 0.7 accuracy in only 5 iterations, while Baseline 2 and 3 require approximately 30 iterations, and Baseline 1 fails to converge. For the more complex Simulation 3, only UUPL and Baseline 3 achieve accuracies exceeding 0.7, with UUPL requiring 32 iterations compared to Baseline 3's 62 iterations. This efficiency is attributed not only to the human uncertainty model but also to the uncertainty-weighted GMM. By synergizing with the entropy-based acquisition function, UUPL selects queries based on human and robot joint uncertainties, ensuring the selected queries are highly informative for the preference learning process. Consequently, our uncertainty-unified framework accelerates convergence rate significantly.
    \item \textbf{Stability}: Finally, we examine the stability of these methods by analyzing the variance of accuracy. UUPL shows a clear trend of decreasing accuracy variance as more data is collected, ultimately achieving low variance, as indicated in Table~\ref{tab}. In contrast, the baseline methods do not exhibit consistent variance reduction. This result highlights that the uncertainty-unified framework produces more stable preference learning outcomes, making the system reliable and applicable in real-world scenarios.
\end{itemize}

Furthermore, we provide the qualitative results by visualizing the learned reward functions for all four methods in the thermal comfort simulation and the tabletop importance simulation, as shown in Appendix~\ref{apdx:2}. The driving choice simulation is omitted, because the 4D feature space makes visualization less practical. Those figures provide a more intuitive understanding of the superior performance of UUPL. 

\vspace{-3pt}
\subsection{Ablation Study}
\label{sec:ablation}
\vspace{-2pt}
In this section, we delve deeper into the individual contributions of the human preference uncertainty model proposed in Section~\ref{sec:uahpm} and the uncertainty-weighted GMM introduced in Section~\ref{uupc}. To avoid potential biases arising from the simplicity of the 1D thermal comfort function, we conduct the evaluation on the more complex and realistic tabletop importance function.

By systematically removing each uncertainty-related component and analyzing their impact, we obtain the results shown in Fig.~\ref{fig:real_results}(a). These findings highlight that both components significantly enhance the overall performance of the system. Specifically, the uncertainty-weighted GMM (yellow curve) optimizes query selection by generating more informative preferences that effectively incorporates human and robot uncertainties, which results in a faster increase in accuracy compared to the baseline (green curve). However, the accuracy variance remains considerable, primarily due to the forced binary choice in uncertain scenarios. Meanwhile, the human preference uncertainty model (purple curve) improves the accuracy of the learned posterior mean by incorporating user-expressed uncertainty levels, and the accuracy variance is significantly reduced by allowing users to describe their preferences with greater precision. Together, these two components synergize to enable the unified uncertainty framework to achieve state-of-the-art performance (red curve).

\vspace{-3pt}
\subsection{User Studies}
\label{sec:us}
\vspace{-2pt}
We conducted three user studies with eight participants to evaluate both our uncertainty-unified framework, especially the user-specific uncertainty calibration process. In this section, we first assessed the calibration process to evaluate its accuracy and reliability in capturing individual uncertainty levels. Following this, we performed two robot experiments with different uncertainty usage to evaluate user experience. The three methods compared were: \citet{Bıyık22024ActivePGP}, referred to as ``no uncertainty'' (NU), which does not include uncertainty modeling; our method without user-specific uncertainty calibration, referred to as ``uncalibrated uncertainty'' (UU); and our full method, referred to as ``calibrated uncertainty'' (CU).

\subsubsection{Calibration Process Evaluation}

We chose the thermal comfort simulation function to be the calibration function $f_{\text{calib}}$, and followed the process introduced in Section~\ref{Sec:usuc}. Fifty preference pairs were collected per participant, and their uncertainty factor values $u^l = \{u^1, u^2, u^3, u^4\}$ were calibrated individually using Algorithm~\ref{code:1}. To evaluate the calibrated uncertainties, we designed a different function $f_t$ and conducted preference learning with all three methods (NU, UU, CU) for all users, suggesting $f_t$ to be their reward function. Each method was repeated three times, with twenty preferences collected per trial. UU and CU experiments were randomized to eliminate potential bias.

The results are presented as the box plot in Fig.~\ref{fig:real_results}(b). NU exhibits the lowest accuracy alongside high variance, indicating the limited effectiveness without uncertainty. While UU achieves higher accuracy due to the inclusion of uncertainty options, it still displays considerable variance, suggesting that participants interpret ``confident'' and ``uncertain'' inconsistently across individuals. In contrast, CU demonstrates both the highest accuracy and significantly reduced variance, validating that UUPL effectively standardizes uncertainty interpretations and better describes the underlying function across participants.
Additionally, user ratings for the learned functions, presented in Fig.~\ref{fig:real_results}(c) as ``User Study 1'', reveal a clear preference for CU ($7.2 \pm 1.0$) over both NU ($5.5 \pm 1.3$) and UU ($5.0 \pm 1.5$). These ratings further reinforce the importance of user-specific uncertainty calibration in achieving more accurate and consistent results.

\subsubsection{Robot Experiments}
\label{sec:robotexp}
To further evaluate the utility of our framework, we designed two real-world experiments with a Kinova Gen 3 manipulator and a Stretch 2 mobile manipulator. These experiments also demonstrate some practical benefits of unified uncertainty.

\begin{figure}[!t]
    \centering
    \includegraphics[width=\columnwidth]{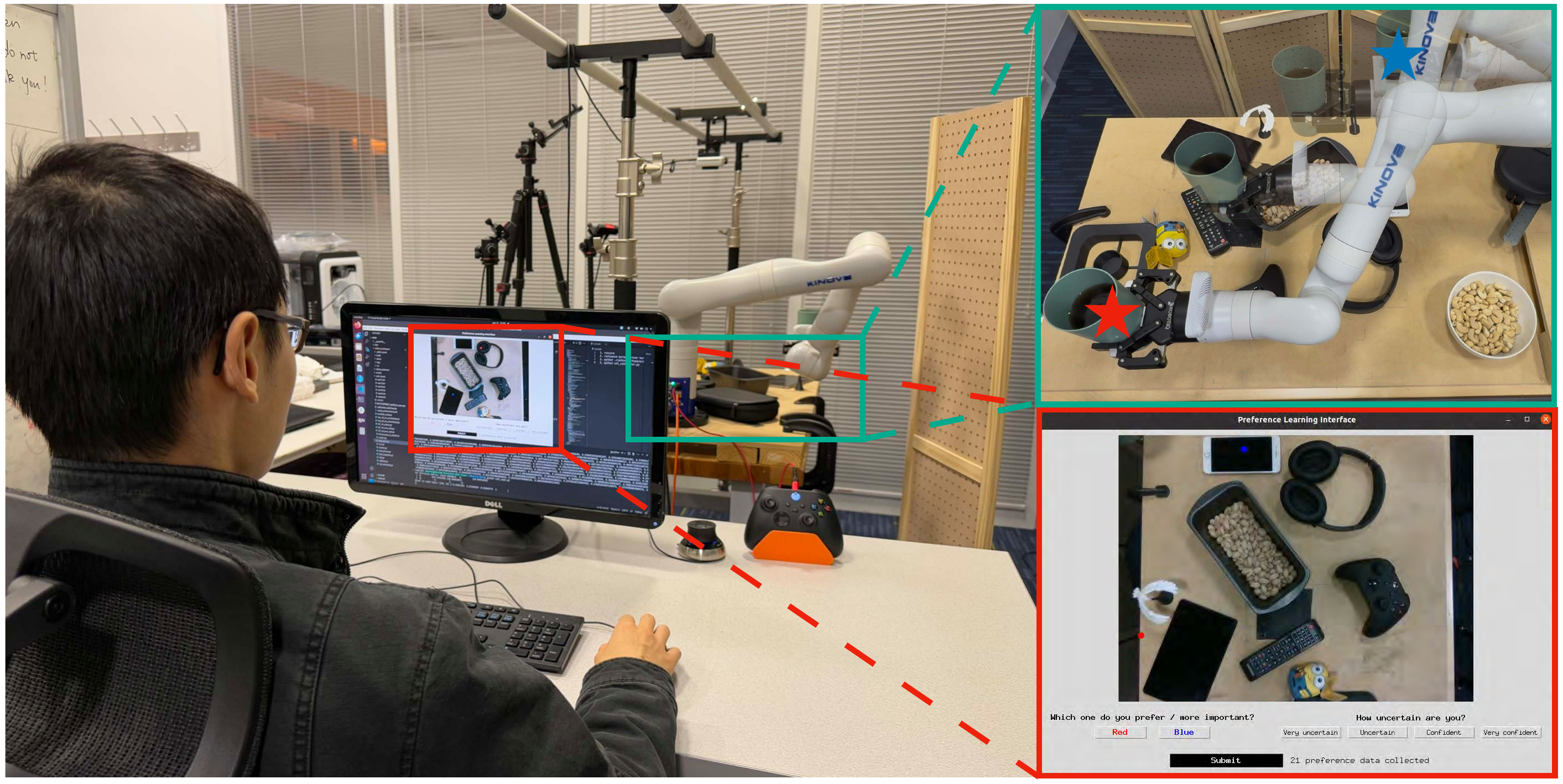}
    \vspace{-15pt}
    \caption{\textbf{One example for the tabletop importance task.} The red box represents the user interface, and green box illustrates the tabletop setup, with the robot trying to move from the blue star to the red star.}
    \label{fig:interface1}
    \vspace{7pt}
\end{figure}

\textbf{Tabletop Importance Task:}
We replicated the tabletop importance simulation task in a real-world setting, where each participant had a unique tabletop configuration. One experiment setup is provided as Fig.~\ref{fig:interface1}. A Kinova Gen 3 was holding a cup of coffee, aiming to transport it to the other side of the table. A top-down image of the tabletop was taken, and participants were instructed to provide preferences and uncertainties for pairs of red and blue points on the image. For a complete set of tabletop configurations, please refer to Appendix~\ref{apdx:3}. During the experiment, each method (NU, UU, CU) was tested three times per participant, with twenty preference pairs collected per trial. Robot trajectories were generated based on the learned reward functions using the A-star planning algorithm. 
In this case, we used the uncertainty to scale the motion velocity: in regions with low variance, the robot moved at a normal speed, while in high-variance regions, it slowed down, mimicking cautious exploration like humans would do in unfamiliar environments.

For each trial, we collected user ratings for both the learned function and the robot’s trajectory (including its uncertainty-scaled velocity). The results are presented in the second and third group columns of Fig.~\ref{fig:real_results}(c) (``User Study 2.1'' and ``User Study 2.2'', respectively). In both cases, CU achieved the highest average human ratings and the lowest overall variance, with ratings of $7.0 \pm 1.1$ for User Study 2.1 and $7.5 \pm 1.0$ for User Study 2.2. In contrast, users did not express a specific preference for NU and UU. These results proved the importance of user-specific uncertainty calibration process, which enables participants to articulate their preferences more precisely, and finally contributes to learning more accurate and user-aligned reward values. Additionally, users perceived the uncertainty-scaled robot motion as being both ``safer'' and ``more natural'', further validating the advantages of incorporating calibrated uncertainty into the UUPL framework.

\begin{figure}[!t]
    \centering
    \includegraphics[width=\columnwidth]{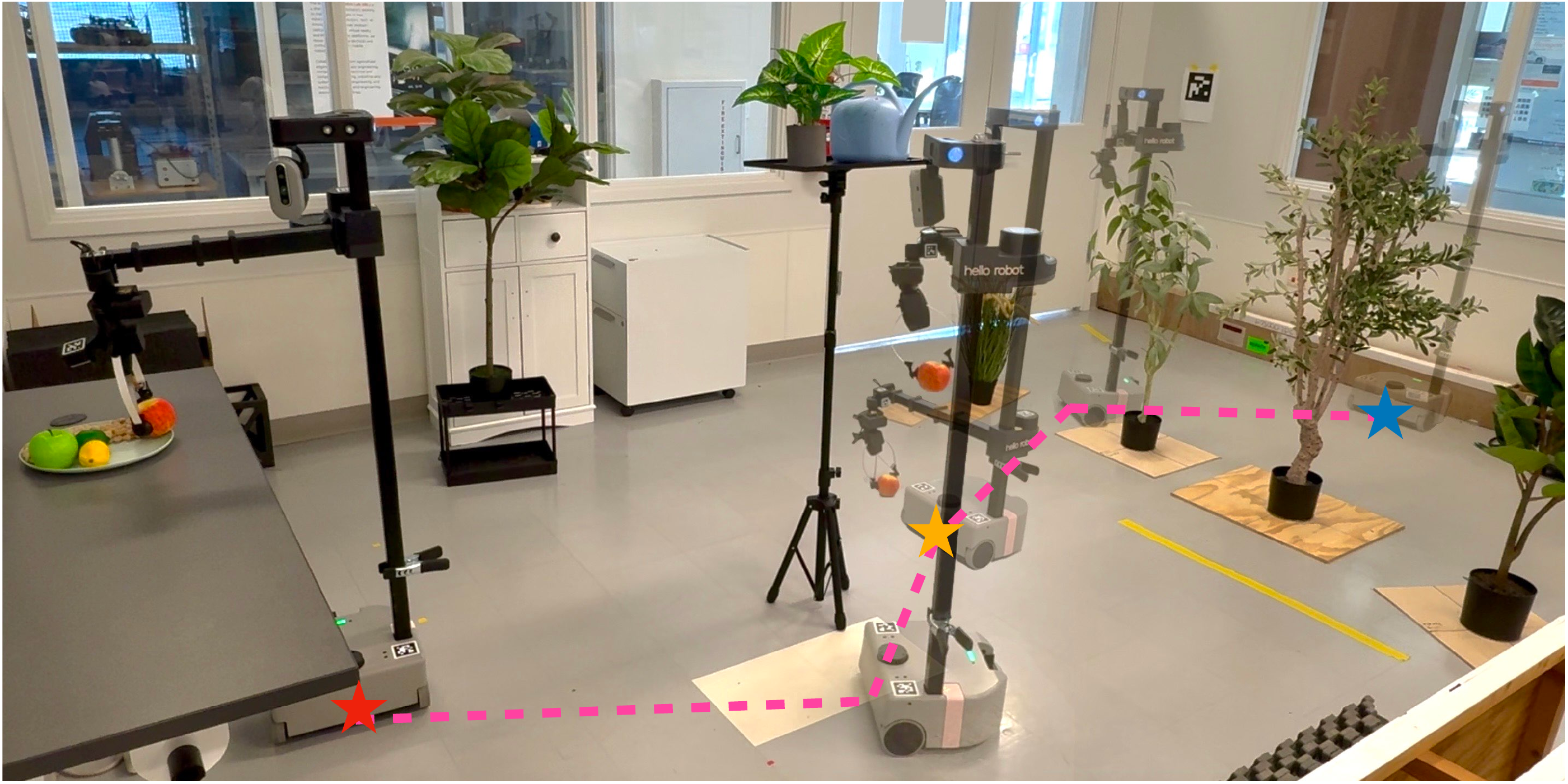}
    \vspace{-15pt}
    \caption{\textbf{Illustration of the apple pick-and-place task.} The dashed pink line is one possible trajectory. The robot starts at the blue star, passes the yellow star, and reaches the red star.}
    \label{fig:exp2}
    \vspace{7pt}
\end{figure}

\textbf{Apple Pick-and-Place Task:}
We further designed an apple pick-and-place experiment using the Stretch 2 mobile manipulator. As shown in Fig.~\ref{fig:exp2}, the robot first traversed through a row of plants to pick up the apple at the yellow star, then it bypassed a rack to the red star and placed the apple on a table. During each trial, users were presented with two trajectories, and their preferences and associated uncertainties were collected.
To encode the trajectories, we extracted three meaningful features: (1) Path through the plants: Whether to take the shortest route, risking the robot becoming stuck or damaged, or follow a safer path at the cost of increased task completion time. (2) Clearance from the rack: The distance maintained while bypassing the rack, balancing the risk of pushing it over with the route efficiency. (3) Traversal velocity: The speed of movement along the path.

As in previous experiments, each method (NU, UU, CU) was tested three times per participant, with twenty preference pairs collected per trial. To address the potential insufficiency of twenty pairs in a higher dimension, we implemented an adaptive query method based on corrected GP variances. Specifically, after collecting twenty pairs, we calculate the overall GP variance. If the variance drop is below than a threshold, indicating a consistent high uncertainty about the learned reward function, five more pairs are collected, repeating until the variance drop reaches the threshold. Since our uncertainty-weighted GMM ensures GP variance reflects a rational and meaningful uncertainty considering both human and robot, UUPL is able to utilize the variance drop as a criterion to stop querying adaptively, which balances model performance and human burden.

We compared NU, UU and CU in this user study from two perspectives: one with a fixed set of twenty preference pairs, and the other one using the adaptive query method, shown as ``User Study 3.1'' and ``User Study 3.2'' respectively in Fig.~\ref{fig:real_results}(c). In the first comparison, UU and CU reached $6.5 \pm 1.2$ and $7.5 \pm 1.3$, both outperforming NU's $5.3 \pm 1.3$. In the second user study, UU and CU improved by approximately $1.0$, but NU remained unchanged. NU's lack of improvement can be attributed to its variance decreasing too rapidly in the absence of uncertainty scaling, causing the threshold to be met consistently after collecting only twenty preference pairs. In contrast, CU effectively regulated the number of queries based on the uncertain extent in users' answers, soliciting additional responses when uncertainty was high. This adaptive query based UUPL achieved the highest rating of $8.6 \pm 0.9$. 

Finally, all users provided an overall average rating of $8.5$ for the uncertainty option design, indicating that it not only facilitates easier and more concise expression of their preferences but also significantly enhances the learning results.

\textbf{Failure Case Analysis:}
We identify several failure cases that may degrade system performance. First, if users consistently select extreme responses (e.g., ``very confident"), the framework effectively reduces to prior methods that ignore uncertainty modeling \cite{Bıyık22024ActivePGP}. Second, when users interpret the notions of ``confident" and ``uncertain" inconsistently across tasks, the Laplace approximation loses accuracy due to fluctuations in the underlying uncertainty levels $u^l$. Finally, if users revise their preference opinions within a single trial, additional noise is introduced into the Gaussian Process model, undermining the learned function.



\section{Conclusion}
\label{sec:conclusion}
In this work, we introduced the concept of uncertainty unification in human-robot interaction (HRI) and proposed uncertainty-unified preference learning (UUPL) within the domain of preference learning. Specifically, we developed a human preference uncertainty model with discrete uncertainty levels, which is integrated seamlessly with the Laplace approximation for Gaussian Process (GP) mean estimation to improve its accuracy. For the GP variance, we proposed an uncertainty-weighted Gaussian Mixture Model (GMM) to unify human uncertainty with robot uncertainty, resulting in interpretable variances that enhance the acquisition function's effectiveness.
We conducted three simulation experiments, three user studies, and an ablation study to evaluate UUPL comprehensively. The results consistently support our claim that unifying human and robot uncertainties improves preference learning performance. Beyond preference learning, we hope the idea of explicit uncertainty unification can inspire more natural HRI in real-world scenarios, while encouraging further research into human-robot joint uncertainty estimation.

\section{Limitation and Future Work}
\label{sec:limitation}
Our work has some limitations that point to opportunities for future research. First, when designing the human preference uncertainty model, we choose four uncertainty level options. While \citet{Wilde2021LearningRF} suggests that humans generally prefer discrete over continuous options, future work is encouraged to determine the optimal number of uncertainty options or adapt this number to individual users. It is worth noting that our framework can adapt to individual preferred granularity by adjusting the number of $u^l$ during calibration without altering the core UUPL framework, ensuring the flexibility and user-friendliness. Besides, it would be beneficial to integrate human uncertainty into the kernel function, which provides more information on representing the similarity of two feature points. Lastly, the weights for uncertainty-weighted GMM are predefined. A promising avenue would be to establish a deterministic relationship between $w^l$ and $u^l$, enabling the GMM to better describe unified uncertainty.

Beyond preference learning, uncertainty unification also has potential applications in other domains. For instance, in developing large language models (LLMs), inferring human uncertainty from linguistic cues such as phrases like ``maybe'', ``perhaps'', or ``I guess'' could be integrated into existing LLM uncertainty models like \cite{Ren2023RobotsTA}. Similarly, in human-robot conversations, implicit signs of uncertainty — such as long pauses, hesitation, or verbal retractions — are also uncertainty-informative and could be utilized to enhance interaction. From a game theory perspective, uncertainty unification could serve as an objective function, where its minima represent the mutual awareness of actions between humans and robots. Finally, a key challenge for future work lies in effectively conveying unified uncertainty to humans, ensuring that it is interpretable and actionable in real-world interactions.

\section*{Acknowledgments}
This work was supported in part by the National Science Foundation under Grant No. 2143435 and AIFARMS through the Agriculture and Food Research Initiative (AFRI) grant no. 2020-67021-32799/project accession no.1024178 from the USDA/NIFA. We would like to thank Dongping Li and Kaiwen Hong for their help in conducting experiments, Neeloy Chakraborty and Shuijing Liu for their valuable feedback during proofreading.


\bibliographystyle{unsrtnat}
\bibliography{references}







\onecolumn

\begin{appendices}
\section{Derivation of Posterior Mean Approximation}
\label{apdx:1}
Let the Laplace-approximated GP mean be $\mathbf{f}_{\text{Lap}}$, then integrate Eq.~\ref{eq:f} and Eq.~\ref{eq:df} into Eq.~\ref{equ:posterior}, we have
\begin{equation*}
\begin{aligned}
    \mathbf{f}_{\text{Lap}} & = \argmax_{\mathbf{f}} P(\mathbf{f} | \mathbf{D})\\
    & = \argmax_{\mathbf{f}} (\ln(P(\mathbf{D} | \mathbf{f})) + \ln(P(\mathbf{f}))) \\
    & = \argmax_{\mathbf{f}} (\ln(\prod_{i=1}^N \Phi \left ( \frac{f(x_i^{(1)})-f(x_i^{(2)})}{u^{l_i}}\right )) \space + (-\frac{N}{2} \ln 2\pi - \frac{1}{2} \ln |\mathbf{K}| - \frac{1}{2}\mathbf{f}^T \mathbf{K}^{-1} \mathbf{f})) \\
    & = \argmax_{\mathbf{f}} (\sum^N_{i=1}\ln{\Phi \left ( \frac{f(x_i^{(1)}) - f(x_i^{(2)})}{u^{l_i}} \right )} - \frac{1}{2} \mathbf{f}^T \mathbf{K}^{-1} \mathbf{f})
\end{aligned}
\end{equation*}
which is exactly Eq.~\ref{eq:pma}.

\vspace{30pt}

\section{Visualization of learned functions}
\label{apdx:2}
Here we provide qualitative results for our simulation experiments.
\begin{figure*}[!h]
    \centering
    \includegraphics[width=\textwidth]{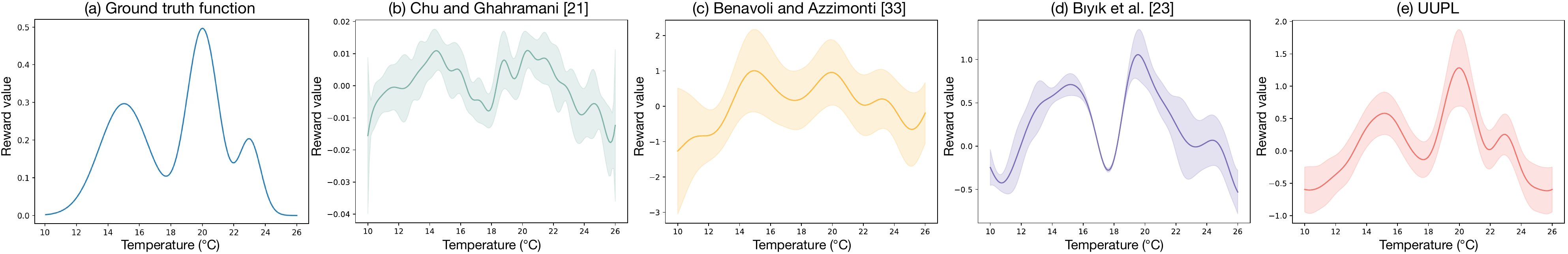}
    \vspace{-10pt}
    \caption{\textbf{Visualization of Sim 1 -- Thermal Comfort.} (a) represents the ground truth function, (b)- (d) illustrate the results obtained using the three baseline methods. (e) presents the outcome of our UUPL approach. Notably, UUPL accurately captures the three local minima along with their relative magnitudes.}
    \label{fig:sim1}
\end{figure*}

\vspace{20pt}

\begin{figure*}[!h]
    \centering
    \includegraphics[width=\textwidth]{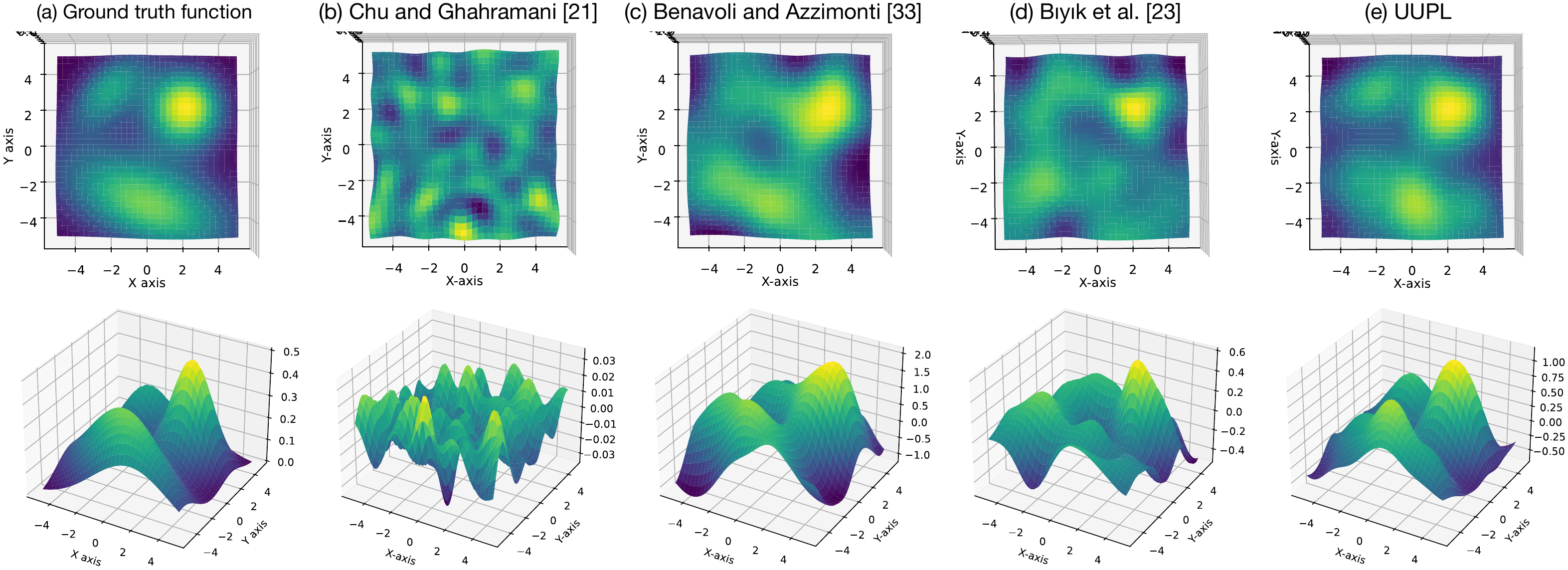}
    \vspace{-10pt}
    \caption{\textbf{Visualization of Sim 2 -- Tabletop Importance.} The first row is the top-down view of the functions in the second row. Column (a) depicts the ground truth function, columns (b)–(d) show the results of the three baseline methods, and column (e) displays the results of UUPL. UUPL successfully identifies the locations and even the shapes of the three objects.}
    \label{fig:sim2}
\end{figure*}

\newpage

\section{Tabletop importance user study configurations}
\label{apdx:3}
\begin{figure*}[!h]
    \centering
    \includegraphics[width=\textwidth]{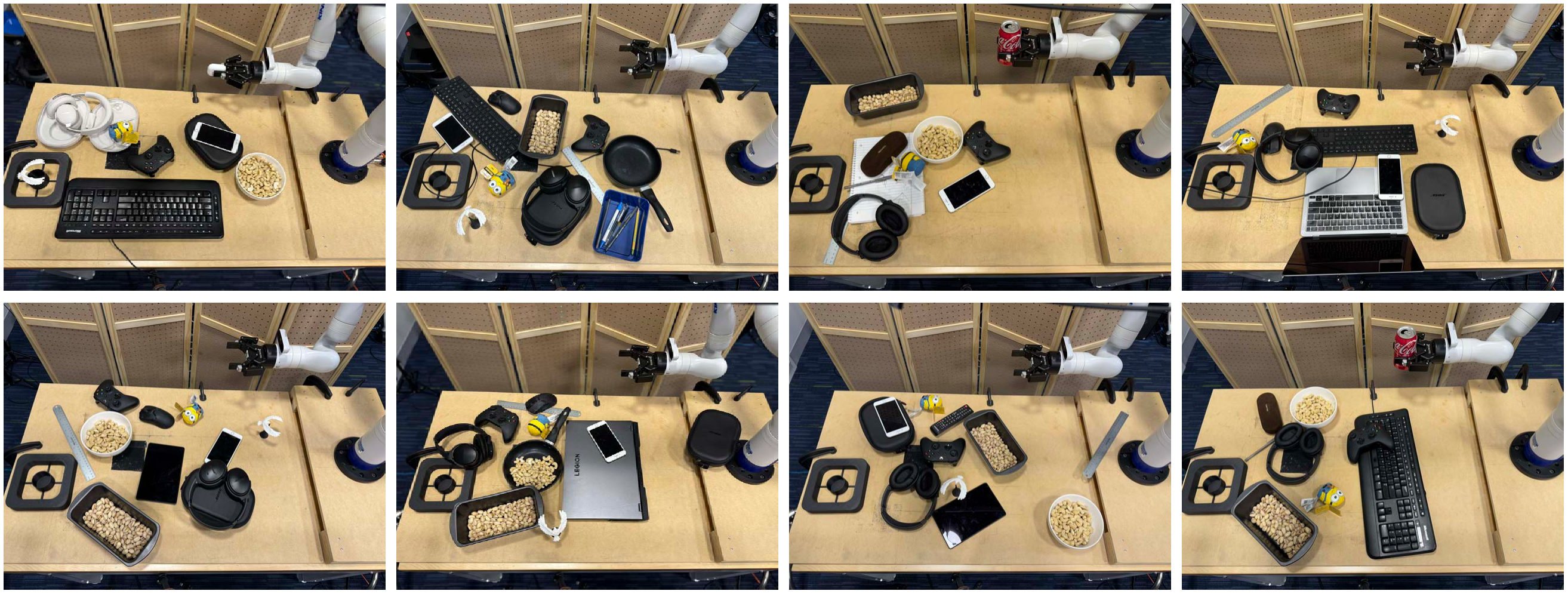}
    \vspace{-20pt}
    \caption{\textbf{The set of configurations for the tabletop importance task.} Each participant was assigned a unique configuration featuring different objects, including electronics, snacks, toys, and stationery.}
    \label{fig:config}
\end{figure*}

\end{appendices}


\end{document}